\documentclass[10pt,twocolumn,letterpaper]{article}

\usepackage[pagenumbers]{cvpr} 

\usepackage{graphicx}
\usepackage{booktabs}
\usepackage{algorithm}
\usepackage{algorithmic}
\usepackage{graphicx}
\usepackage{mathptmx}
\usepackage{amsmath}
\usepackage{amssymb}
\usepackage{amsthm}

\newtheorem{thm}{Theorem}[section]
\newtheorem{lem}[thm]{Lemma}

\newtheorem{cor}{Corollary}

\theoremstyle{definition}

\usepackage[pagebackref,breaklinks,colorlinks]{hyperref}

\usepackage[capitalize]{cleveref}
\crefname{section}{Sec.}{Secs.}
\Crefname{section}{Section}{Sections}
\Crefname{table}{Table}{Tables}
\crefname{table}{Tab.}{Tabs.}

\newcommand{\R}{\mathbb{R}}
\newcommand{\Z}{\mathbb{Z}}
\newcommand{\bS}{\mathbb{S}}

\newcommand{\bv}{\mathbf{v}}
\newcommand{\F}{\mathcal{F}}
\newcommand{\x}{\mathbf{x}}
\newcommand{\ECT}{\operatorname{ECT}}
\newcommand{\EC}{\operatorname{EC}}

\begin{document}

\title{Euler Characteristic Transform Based Topological Loss for Reconstructing 3D Images from Single 2D  Slices\textbf{}}

\author{Kalyan Varma Nadimpalli\\
IIIT Bangalore\\
{\tt\small Kalyan.Varma@iiitb.ac.in}
\and
Amit Chattopadhyay\\
IIIT Bangalore\\
{\tt\small a.chattopadhyay@iiitb.ac.in}
\and
Bastian Rieck\\
Helmholtz Munich\\
{\tt\small bastian.rieck@helmholtz-muenchen.de	}
}
\maketitle

\begin{abstract}
The computer vision task of reconstructing 3D images, i.e., shapes, from their single 2D image slices is extremely challenging, more so in the regime of limited data. 
Deep learning models typically optimize geometric loss functions, which may lead to poor reconstructions as they ignore the structural properties of the shape.  
To tackle this, we propose a novel topological loss function based on the \emph{Euler Characteristic Transform}. This loss can be used as an inductive bias to aid the optimization of any neural network toward better reconstructions in the regime of limited data. We show the effectiveness of the proposed loss function by incorporating it into SHAPR, a state-of-the-art shape reconstruction model, and test it on two benchmark datasets, viz., Red Blood Cells and Nuclei datasets. We also show a favourable property, namely injectivity and discuss the stability of the topological loss function based on the Euler Characteristic Transform.
\end{abstract}

\section{Introduction}
\label{sec:intro}
Our brains possess the amazing ability to be able to reconstruct 3D shapes from  single 2D images by leveraging prior knowledge and inductive biases about the shapes and sizes of objects based on the information captured from previously observed objects \cite{3DRefPaper}. However, for a computer this inverse problem is ill-posed and extremely challenging. This is because for a single 2D image the space of possible 3D reconstructions is very large and often ambiguous.

There have been prior deep learning-based attempts to solve this challenge, but most of them rely on large datasets and/or 3D models of the shape \cite{chang2015shapenet, sun2018pix3d, kolotouros2019convolutional}. The biomedical setting in which we consider this problem, unfortunately, does not provide large labeled datasets and it is too expensive to construct them. The sizes of the datasets available in the biomedical domain are orders of magnitude smaller than the ones available in other domains. To this end, we focus on improving reconstruction performance not by using 3D models or large datasets but instead by adding additional inductive biases in the form of a topology-based regularization to the optimization process. 
Most models typically optimize geometry-based loss functions that work on a per-pixel basis, such as  the DICE loss. We improve the performance of an existing neural network by adding a novel complementary topology-based loss that considers more global topological features, such as connectivity, tunnels, or voids. Specifically, we design a novel regularization term based on the Euler Characteristic Transform~\cite{turner2014persistent}, that is 
computationally efficient, can work with any image size and can be plugged into any neural network. An overview of how our loss function can be used can be seen in Fig \ref{archFig}.
We demonstrate the efficacy of the proposed loss function by plugging it into the SHAPR model and testing it on two bio-medical datasets used in the prior work \cite{Shapr, topo_shapr}. In the current paper, our key contributions   are as follows:
\begin{itemize}
    \item We adapt the Euler Characteristic Transform (ECT), 
    obtaining a novel topological loss function for 3D shape reconstruction
    that is compatible with any neural network architecture.
    
    \item We prove conditions for our proposed ECT-based loss to be injective as well as discuss stability results of ECT on binary images.  
    
    \item We show the effectiveness of the proposed method by training the SHAPR model\cite{Shapr} with our proposed loss on two benchmark datasets. We see significant improvements in almost all metrics compared to the prior work on both datasets.
\end{itemize}

\noindent
\textbf{Outline.} 
In Section \ref{sec:prior-work} we go over the literature relevant to our work. In Section \ref{sec:math-back} we then briefly explain the mathematical background required to understand our work. Subsequently, in Section \ref{sec:method} we describe our proposed loss functions in detail as well as how they fit into the overall training of a neural network. In Section \ref{sec:theory} we then prove and discuss favourable mathematical properties of the ECT. In Section \ref{sec:Experiments} we then demonstrate the efficacy of our model and discuss the significance of our results. Finally in Section \ref{sec:conclusion} we summarize our work and list some potential future work.

\section{Prior Work}
\label{sec:prior-work}
Multiple variants of the problem of 2D to 3D image reconstruction have been studied by various communities for different applications like scene understanding, 
medical, robot navigation, etc. \cite{3DRefPaper}. The tasks considered differ in their input type, some variants consider multiple slices as the input while some consider a single image like in our formulation. Among the models that only take a single image as an input, most of them require a synthetic 3D model of the output or very large datasets \cite{chang2015shapenet, sun2018pix3d, kolotouros2019convolutional}.

The application of computational topology to machine learning is an emerging field that has shown promise in various applications \cite{hensel2021survey}. It has recently been used extensively in computer vision tasks like segmentation, image generation, etc. \cite{hu2021topology, thresholding, wang2020topogan}. 
In the current paper, we improve the performance of image reconstruction models using tools from topology, namely, the Euler Characteristic Transform \cite{turner2014persistent}.

SHAPR \cite{Shapr} is the first machine learning model that considers the problem of 2D to 3D reconstruction in the case of biomedical images. This model proved to be significantly better than standard synthetic models like a cylindrical fit and ellipsoid fit. They also showed that features extracted from the 3D reconstruction helped to improve accuracy in downstream classification tasks on the 2D images.
Recently, a diffusion-based model DISPR\cite{waibel2022diffusion} has been introduced that outperforms the GAN-based SHAPR model.

Waibel \etal~\cite{topo_shapr} extend the SHAPR model by training the model on a combined loss function of both the DICE loss as well as a regularization term defined by the Wasserstein distance between the persistence diagrams---topological descriptors---of the predicted shape and the ground truth. This model outperforms the SHAPR model and provides much better reconstructions than the vanilla SHAPR model. However, it has been shown by Oner \etal~\cite{thresholding} that such persistence diagram based loss functions are not optimal for the following reasons:
\begin{itemize}
    \item Since the ground truth images are binary images, calculating the persistence diagrams over the filtration of pixel values degenerates to calculating the Betti numbers, which is a topological measure of limited expressivity.
    
    \item Persistence diagrams throw away location information and are generally not injective mappings, thus potentially leading to erroneous matchings, which in turn may lead to wrong reconstructions.
\end{itemize}
To overcome these drawbacks, we exploit the injectivity property and low computational cost of the ECT, developing a novel ECT-based loss function that is more expressive and serves as a better optimization term for a neural network.




\begin{figure*}

\begin{subfigure}{0.25\textwidth}
  \centering
  \includegraphics[width=\textwidth]{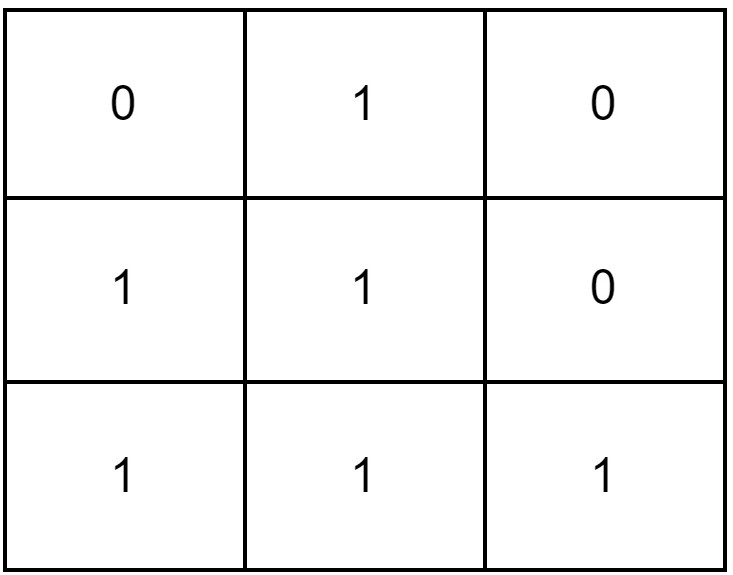}    
  \caption{}
    \label{CC_a}
  
\end{subfigure}%
\hfill
\begin{subfigure}{0.20\textwidth}
  \centering
  \includegraphics[width=\textwidth]{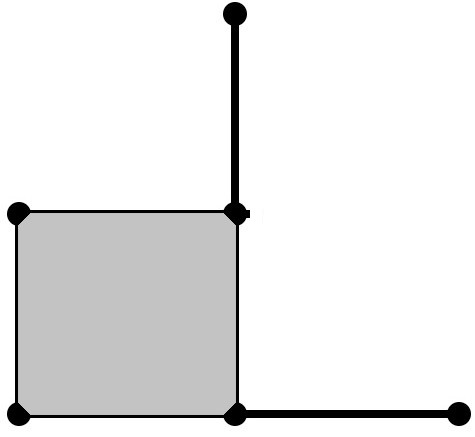}    
  \caption{}
  \label{CC_b}
\end{subfigure}%
\hfill
\begin{subfigure}{0.25\textwidth}
  \centering
  \includegraphics[width=\textwidth]{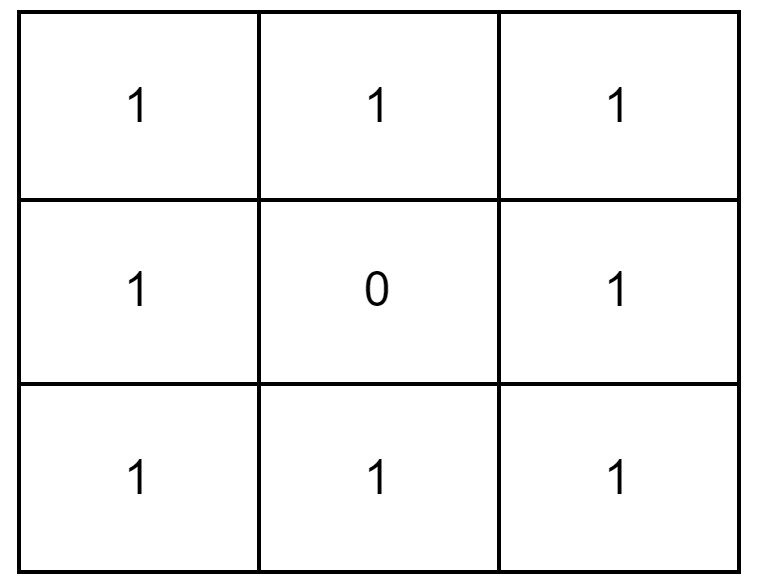}    
  \caption{}
  \label{CC_c}
\end{subfigure}%
\hfill
\begin{subfigure}{0.20\textwidth}
  \centering
  \includegraphics[width=\textwidth]{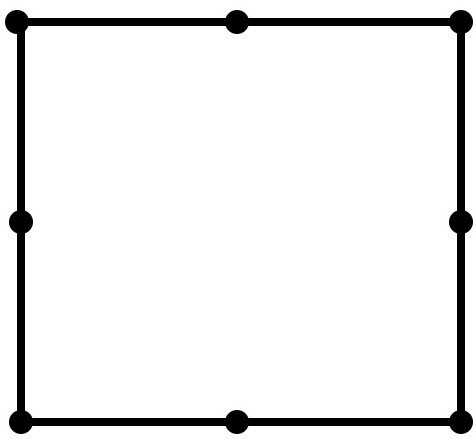}    
  \caption{}
  \label{CC_d}
\end{subfigure}%
\caption{Examples of cubical complex construction from binary images. (b) and (d) are the cubical complexes corresponding to the binary images (a) and (c), respectively.}
\label{img:CC_Construction_Image}
\end{figure*}

\section{Mathematical Background}
\label{sec:math-back}
In this section, we briefly introduce the mathematical background required for our work, for a more detailed explanation we refer the reader to Edelsbrunner \etal and Turner \etal~\cite{harer, turner2014persistent}.

\subsection{Simplicial and Cubical Complex}
\label{subsec:CubicalComplex}
A simplicial complex is the fundamental building block of algebraic topology, comprised of simplices. A $k$-simplex $\sigma$ can be understood as the convex hull of $k+1$ affinely independent points. A $0$-simplex is a point, a $1$-simplex is an edge, a $2$-simplex is a triangle and a $3$-simplex is a tetrahedron. 
A face $\tau$ of a simplex is the convex hull of a subset of the $k+1$ points. It is often represented as a face by the notation $\tau \preccurlyeq \sigma$. A simplicial complex $K$ is a finite collection of simplices satisfying two conditions: 
\begin{enumerate}
    \item $\sigma \in K$ and $\tau \preccurlyeq \sigma$ implies that $\tau \in K$
    \item $\sigma, \sigma_0 \in K$ implies $\sigma \cap \sigma_0$ is either empty or a face of both. 
\end{enumerate} 

The dimension of the simplicial complex is the dimension of the largest simplex in the complex, denoted by Dim($K$). A subcomplex $L$ is a subset of a simplicial complex $K$. 
$K^d$ is a particular subcomplex that is defined as a subcomplex consisting of all simplices of dimension $d$ from $K$, that is, $K^d = \{\sigma \in K \mid \text{dim}(\sigma) = d\}$.

A cubical complex is a special variant of a simplicial complex that is particularly useful in representing grid-like shapes. It has recently caught traction in applications for image processing due to the fact that it is better aligned to the grid-like structure of images \cite{fmri, allili2001cubical}.
Informally, a cubical complex is identical to a simplicial complex except the $n$-simplices are replaced with $n$-cubes. For example, the triangles ($2$-simplices) are replaced by squares ($2$-cubes), and tetrahedra ($3$-simplices) by cubes ($3$-cubes) and so on. 

Given a $d$-dimensional binary image, a natural way to convert it to a cubical complex is by defining the $0$-cubes as the set of voxels. Then an $i$-dimensional cube is formed by connecting a set of $2^i$ adjacent voxels whose voxel values are $1$. Note that two $d$-dimensional voxels are adjacent if they share a $(d-1)$-dimensional face. Thus $1$-cubes are the edges corresponding to two adjacent voxels with values $1$. Similarly, the $2$-cubes are the squares corresponding to  four adjacent voxels with values $1$ and so on. 
We illustrate this construction by example in Figure \ref{img:CC_Construction_Image}. Figures \ref{CC_b} and \ref{CC_d} are is the cubical complexes corresponding to Figures \ref{CC_a}, \ref{CC_c} when converted by the above procedure.

\subsection{Sublevel Sets and Filtrations}
Consider a simplicial or cubical complex $K$ and a monotonic function $f: K \rightarrow \R$. By $f$ being monotonic, we mean $f(\sigma) \leq f(\tau)$ whenever $\sigma \preccurlyeq \tau$.
For such  monotonic functions, the sublevel set $K(a)$ corresponding to a real value $a$ is defined by
    \begin{equation*}
        K(a) = f^{-1}(-\infty, a],
    \end{equation*} 
    which is a subcomplex of $K$. If there are $m$ simplices in $K$, as we increase $a$, we get $n+1$ $\le m+1$ different subcomplexes which we can be arranged in an increasing sequence,
    \begin{equation*}
        \emptyset = K_{0} \subseteq K_{1} \ldots \subseteq K_{n} = K 
    \end{equation*}
    where $K_{i} = K(a_{i})$ and  $a_{1} < a_{2} < \ldots < a_{n}$ are the distinct function values of $f$ at the vertices of the simplicial complex $K$.
    This sequence of complexes is called the filtration of $K$ with respect to $f$.
    A common filtration we consider is the height filtration. Given a height $h$ and a particular direction $\Vec{u}$, we define the sub-complex $K_{\vec{u}, h}$ consists of all the simplices of $K$ whose vertices have height $\le h$ along the direction $\vec{u}$. We can naturally define a filtration by increasing the value of $h$ along the direction $\vec{u}$.

\subsection{Euler Characteristic Curve}
Given a simplicial complex $K$, the Euler Characteristic Curve of $K$ along a direction $\vec{u}$ is a function $EC_{\vec{u}, K}: \R \rightarrow \Z$ defined by 
\begin{equation}
\label{eqn:EC}
h\mapsto \chi(K_{\vec{u}, h}), 
\end{equation}
where $\chi(K_{\vec{u}, h})$ is the Euler characteristic of the simplicial complex $K_{\vec{u}, h}$, which is defined as
\begin{equation}
\chi(K_{\vec{u}, h})=\sum_{i=0}^{d} (-1)^d \mathrm{Card}\;K_{\vec{u}, h}^i,
\end{equation}
where $\mathrm{Card}(K_{\vec{u}, h}^i)$ denotes the number of $i$-simplices in the subcomplex $K_{\vec{u}, h}^i$.
By computing the Euler characteristic alongside a filtration, we obtain the \emph{Euler Characteristic Curve}. This construction works for general filtrations and is not restricted to the height filtration.

\subsection{Euler Characteristic Transform}
The Euler Characteristic Transform (ECT)\cite{turner2014persistent} of a $d$-dimensional simplicial complex $K$, denoted by $\ECT_K: \bS^{d-1} \xrightarrow[]{} \Z^{\R}$, is defined by
\begin{equation}
    \vec{v} \xrightarrow[]{} \EC_{\vec{v}, K},
\end{equation} 
where the direction  $\vec{v}$ is chosen from the $(d-1)$-dimensional unit sphere $\bS^{d-1}$. That is, the ECT is the set of all Euler Characteristic Curves obtained over the height filtrations along all possible directions. 
The ECT is the heart of our method. We use it as a topological descriptor to capture the important topological features of 3D images to define our topological loss functions.

\begin{figure*}
    \centering
    \includegraphics[width=0.91\textwidth]{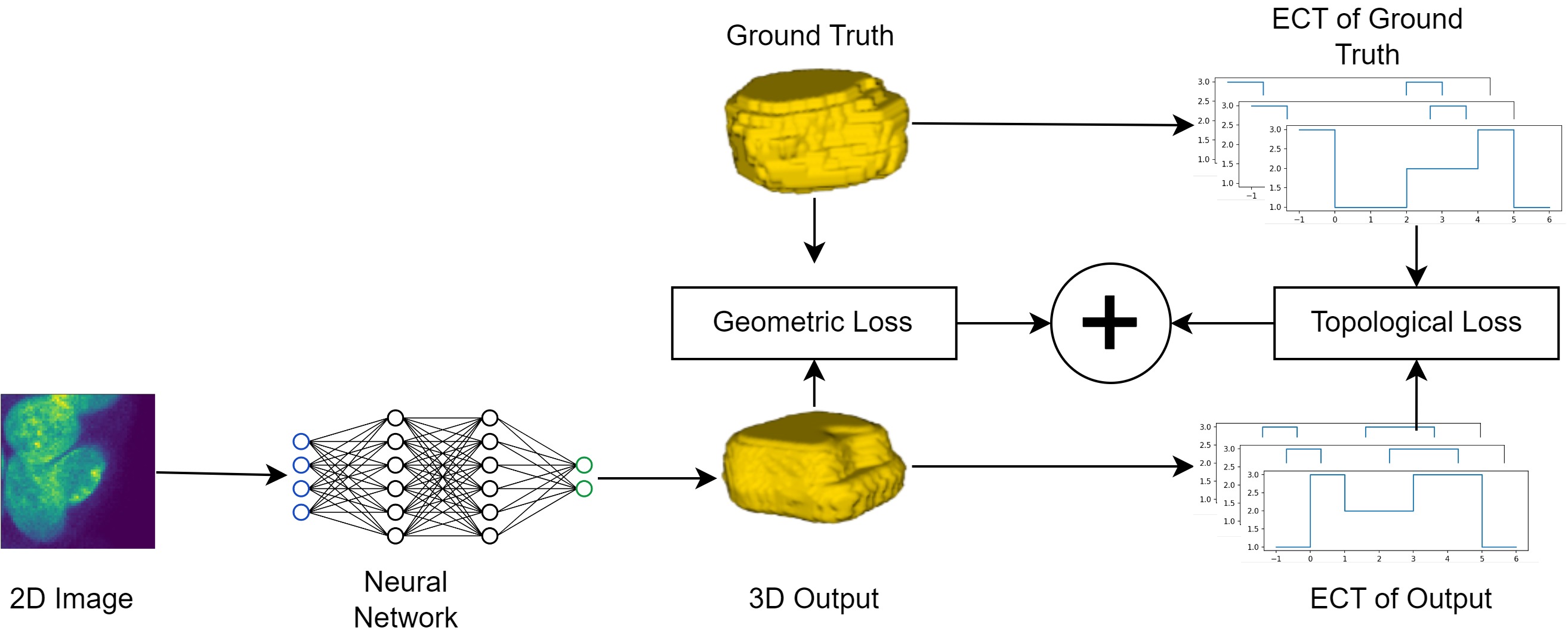}
    \caption{Workflow of our proposed method. Given a 2D image, a neural network produces a 3D output. The neural network is then trained on the sum of a geometric loss function like DICE loss or BCE and our proposed topological loss function, the distance between the ECTs of the images. Neural network image generated from \cite{lenail2019nn}.}
    \label{archFig}
\end{figure*}

\paragraph{Distance between ECTs.}
The distance between two  ECTs corresponding to two complexes $K_1$ and $K_2$ is defined by
\begin{equation}
\label{eqn:distance}
d(\ECT_{ K_1}, \ECT_{ K_2}) = 
\displaystyle\int_{\vec{u}\in \bS^{d-1}} \|\EC_{\vec{u}, K_1} - \EC_{\vec{u}, K_2}\| ^ {2}\mathrm{d}u,
\end{equation}
where $\|.\|$ is the $l_2$-norm.  We use this distance to compute the topology based loss function to train our neural network. In practice, the integration in equation~(\ref{eqn:distance}) is computed using the Monte Carlo method, i.e., we compute the average of the $l_2$-norms between the Euler curves along a finite number randomly sampled  directions from $\bS^{d-1}$.

\section{Our Method: Euler Characteristic Transform-based Loss}
\label{sec:method} 
In this section, first we describe the overall workflow of our 3D image reconstruction method and how our loss function fits into a neural network training procedure (subsection \ref{subsec:overall_model}). Subsequently, we give the detailed algorithm  to compute the proposed loss functions (subsections \ref{subsec:loss-term}, \ref{subsec:euler_curve_computation}).

\subsection{Overview}
\label{subsec:overall_model}
In our method, we develop a loss function based on the ECT  to train a neural network for 3D image reconstruction from a single 2D slice.
Figure \ref{archFig} shows the workflow of our model, which is explained in the following steps.
\begin{enumerate}
    \item Given a 2D slice, it is first passed through a neural network that gives an output 3D image $I$, where for each voxel $x$ of $I$, the model assigns the likelihood of $x$ being part of the true 3D image.

    \item Given the 3D prediction $I$ and the 3D ground truth $Y$, we use the DICE loss combined with a scaled ECT-based loss function, denoted  $L_{TL}$, to optimize the neural network. Mathematically, this can be represented as:
    \begin{equation}
    L(I, Y) = L_{DICE}(I, Y) + \lambda\;L_{TL}(I, Y ),
    \end{equation}
    where $\lambda$ is the weight parameter for the topological loss term.
\end{enumerate}
This is a similar setup as described by Waibel \etal \cite{topo_shapr}, however, our method differs in the details of the topology-based loss  $L_{TL}(I, Y)$ and as a result the efficacy of it as well. 
Next, we discuss our algorithm to train a neural network by computing the topological loss terms based on ECT, in detail.

\subsection{ECT Based Training Algorithm}
\label{subsec:loss-term}
Given a dataset of 2D slices and corresponding 3D images, we first train the SHAPR model based on the proposed ECT-based loss function. 
For ease of understanding, in 
\begin{algorithm}[H]
    \caption{\sc{TrainSHAPRModelBasedOnECT}}
    \label{algo:SHAPRAlgo}
\textbf{Input:} $X$ - \textit{2D image slice},
\\
\hspace*{1cm} $Y$ - \textit{Corresponding 3D ground-truth image,}\\  
\hspace*{1.1cm}$n$ - \textit{Number of thresholds,} \\
\hspace*{1cm} $\Theta_0$ - \textit{Initial model parameters}\\
\textbf{Output: } \textit{Trained SHAPR Model}
\begin{algorithmic}[1]
\STATE $\Theta \xleftarrow[]{} \Theta_0$ \hfill \% \textit{Initialize  model parameters}\\[1.2ex]
    \FOR{epoch $= 1, 2, \ldots, N$}        
        \STATE $I \xleftarrow[]{}$ SHAPR($X$, $\Theta$)\\
        \STATE $L_{Topo} \xleftarrow[]{} 0$ \hfill  \% \textit{Initialize the Topology Loss}\\[1.2ex]
        \STATE \% \textit{Unfold and Sort the Distinct Voxel Values of $I$ and $Y$ in Array $R$}
        \STATE $R \xleftarrow[]{} $SortDistinct~($I\cup Y$)
        \STATE $m \xleftarrow[]{} length(R)$ \hfill \%\textit{Number of Voxels in $I$}\\[1.2ex]
        \STATE \% \textit{Compute ECT-based Topological Loss}
        \FOR{thresh = $R([\frac{m}{n}]),\, R([\frac{2m}{n}]),\,\ldots,\, R([m])$}
            \STATE $A \xleftarrow[]{} $BinaryImg($I$, thresh)
            
            \STATE $B \xleftarrow[]{} $BinaryImg($Y$, thresh)
            \STATE $\ECT_A \xleftarrow[]{} $ComputeECT($A$)
            \STATE $\ECT_B \xleftarrow[]{}$ ComputeECT($B$)
            \STATE $L_{Topo}$ += $(\ECT_A - \ECT_B)^{2}$
        \ENDFOR\\[1.2ex]
        \STATE \% \textit{Compute Total Loss}
        \STATE $L = L_{DICE} + \lambda L_{Topo}/n$\\[1.2ex]
        \STATE \% \textit{Perform Gradient Update Step to Update the Model Parameters $\Theta$ with Learning Rate $\alpha$}
        \STATE $\Theta \xleftarrow[]{} \Theta - \alpha \,\nabla_{\Theta}L$ 
    \ENDFOR
\end{algorithmic}
\end{algorithm}
\noindent
Algorithm~\ref{algo:SHAPRAlgo}, we demonstrate the training of the SHAPR model on a single training sample, i.e., using a 2D slice image $X$ and its ground-truth 3D image $Y$.  
In every epoch (or training step), the image $X$ is first passed to the  SHAPR model and an output 3D image $I$ is produced by the model (Line $3$). Then at each step, we compute the ECT-based loss
    function using Monte Carlo sampling. For this, we sort the distinct voxel values of $I$ and $Y$ in a one-dimensional array $R$ and find $n$ equally spaced thresholds $R([\frac{m}{n}]),\, R([\frac{2m}{n}]),\, \ldots,\, R([m])$ where $m$ is the total number of  voxels in $I$ (Lines $5$-$9$). For each threshold~$\tau$, we compute the binary images $A=\mathbb{I}[x\geq \tau \mid x\in I]$ and $B=\mathbb{I}[x\geq \tau \mid x\in Y]$ corresponding to $I$ and $Y$, respectively, where $\mathbb{I}$ is the indicator function (Lines $10$-$11$). Next, for each of these binary images $A$ and $B$, we compute the Euler Characteristic Transforms $ECT_A$ and $ECT_B$ using Algorithm~\ref{algo:ECTAlgo} (Lines $13$-$14$). Finally, the topological loss function is computed using the average of $l_2$-norms between $ECT_A$ and $ECT_B$, for all thresholds.
A scaled version of this topological loss (here, $\lambda$ is the scaling factor) is added with the standard DICE loss to compute the total loss (Line $17$). The model parameters $\Theta$ are then updated by optimizing this loss using a gradient descent method (Line $19$). Note that we choose equally spaced thresholds on the sorted array $R$  to obtain a more varied set of images as compared to thresholding based on the range of voxel values.

\subsection{ECT Computation}
\label{subsec:euler_curve_computation}
In this sub-section, we explain the details of computing the Euler Characteristic Transform for a binary image $A$ whose pseudocode is given in Algorithm \ref{algo:ECTAlgo}. 
Broadly, we first construct a cubical complex $C$ from the binary image $A$ using the method explained in Section \ref{subsec:CubicalComplex} (Line 1, Algorithm \ref{algo:ECTAlgo}). Then to approximate the ECT we sample $l$ random directions from the unit sphere $\bS^2$ (where $l$ is a chosen parameter), and for each sampled direction $u$ we compute the Euler Curve of $C$ along the direction $u$ (Lines 4-6). The obtained set of $l$ Euler Curves is returned as our Euler Characteristic Transform (Line 8).

\begin{algorithm}[H]
    \caption{\sc{ComputeECT}}
    \label{algo:ECTAlgo}
\textbf{Input}: $A$ - 3D Binary Image\\
\textbf{Output}: $ECT_A$ 
\begin{algorithmic}[1]
    \STATE $C \xleftarrow[]{}$ CubicalComplex(A) 
    \STATE $\ECT_A \xleftarrow[]{} []$ \hfill \% \textit{Initialize as an empty array}\\[1.2ex]
    \STATE \% \textit{Compute Euler curves along $l$ sampled directions chosen from the unit sphere $\bS^2$}
    \FOR{$i = 1, 2, \ldots, l$}
        \STATE $\vec{u} \xleftarrow[]{}$ sampleDirection($\bS^2$)
        \STATE $\ECT_A$.add(EulerCurve($C, \vec{u}$))
    \ENDFOR\\[1.2ex]
    \RETURN $\ECT_A$
\end{algorithmic}
\end{algorithm}

The Euler Curve computation of a cubical complex $C$ along a sampled direction $\vec{u}$ is described in Algorithm \ref{algo:EulerCurve}. We compute the minimum $h_{min}$ and maximum $h_{max}$ of all heights of the vertices in the cubical complex $C$ along the direction $\vec{u}$ (Lines 1-2). For a chosen parameter $M$, we sample the height field at $M+1$ equally spaced heights of step-size $dh$ (Line 5).
For each sampled height, we calculate the Euler Characteristic of the sub-complex $C_{\vec{u}, h}$ (Lines 7-10). We return the list of obtained values as our discrete representation of the Euler curve. Note that the smaller the step size $dh$, the closer our representation is to the continuous Euler Curve. 

\begin{algorithm}[H]
    \caption{\sc{EulerCurve}}
    \label{algo:EulerCurve}
\textbf{Input:} $C$ - Cubical complex , $\vec{u}$ - Direction vector \\
\textbf{Output:} $\EC_{u, C}$
\begin{algorithmic}[1]
    \STATE $h_{min} \leftarrow \min({\vec{u}\cdot \bv_0,\ldots, \vec{u}\cdot \bv_n})$  
    \STATE $h_{max} \leftarrow \max({\vec{u}\cdot \bv_0,\ldots, \vec{u}\cdot \bv_n})$  
    \STATE $\EC_{u, C} = []$ \hfill \%\textit{Initialize as an empty array}
    \STATE $h \leftarrow  h_{min}$
    \STATE  $dh=(h_{max}-h_{min})/M$ \hfill \% \textit{Step length with parameter $M$}\\[1.2ex]

    \STATE \% \textit{Compute Euler curve of $M+1$ steps}
    \WHILE{$h \leq h_{max}$}
        \STATE $\EC_{u, C}.\text{add}(\kappa(C_{\vec{u}, h}))$ 
        \STATE $h \mathrel{+}= dh$
    \ENDWHILE\\[1.2ex]
    \RETURN $\EC_{u, C}$
\end{algorithmic}
\end{algorithm}

\section{Theoretical Properties}
\label{sec:theory}
In this section, we analyze and prove some important properties of the transform and the proposed loss function to evaluate our method.

\subsection{Injectivity Property}
\label{subsec:injectivity}
Turner \etal \cite{turner2014persistent} have shown that ECT over the space of simplicial complexes in $\R^3$ is injective. In our method, for computing the ECT-based loss function between two 3D images, we obtain a sequence of binary images corresponding to a sequence of threshold values for each 3D image and compute the ECT for each of the binary images (Algorithm \ref{algo:SHAPRAlgo}, Line 9-13). In the following lemma, we provide a criterion to choose the number of thresholds so that such sequences of binary images are different for two different 3D images. Subsequently, sequences of ECTs for two different 3D images will also be different. Let, $\F(I, t)$ denote the sequence of $t$ binary images from the 3D image $I$, for $t$ equally spaced threshold values. Then we have the following result.
\begin{lem}
    Given two 3D images $I_1$ and $I_2$, $\F(I_1, t) = \F(I_2, t)$ iff $I_1 = I_2$, provided $t \geq $ distinct number of voxel values in  $I_1\cup I_2$. 
\end{lem}
\begin{proof}
    We prove if $I_1 \neq I_2$ then $\F(I_1, t) \neq \F(I_2, t)$. 
    Since $t \geq $ distinct number of voxel values in  $I_1\cup I_2$, we threshold on every distinct value in $I_1\cup I_2$. As $I_1 \neq I_2$, at some coordinate $x$, $I_1(x) \neq I_2(x)$. Without loss of generality, let $I_1(x) < I_2(x)$. Now when we threshold at value $I_2(x)$, $I_1(x)$ becomes $0$, while $I_2(x)$ is $1$. This implies that $\F(I_1, t) \neq \F(I_2, t)$, as at the image thresholded at $I_2(x)$, we obtain different images. 
\end{proof}

We note, the distance function between two ECTs, in equation (\ref{eqn:distance}),  satisfies the metric property \cite{WECT, turner2014persistent}. We use this in the subsequent proofs. 

\subsection{A Discussion on Stability of ECT-Based Loss}
A commonly studied property in computational topology is the stability of a transform, that is the effect of perturbations on the input to the transformed output \cite{cohen2005stability, skraba2020wasserstein}. We discuss a similar property for the case of ECT on binary images. We bound the possible change in the ECT of a binary image by a constant proportional to the size of the image and the size of the change in the input. We first prove a necessary lemma for our proof in Lemma \ref{lem:cubeCount}. We then show that the distance between two EC's is bounded in Theorems \ref{thm: EC_1} and \ref{thm:EC_full}. Subsequently we prove that the ECT is bounded in Corrollary \ref{cor:ECT}. We then discuss the effect of thresholding on stability.

    \begin{lem}
        A vertex in a $d$-dimensional grid is a part of at most $3^d$ cubes of any dimension.
    \label{lem:cubeCount}
        
    \end{lem}
    \begin{proof}
        Consider a vertex $\bv_0=(x_1, x_2, \ldots, x_d)$  in the interior of the grid. Every $k$-cube, that has $\bv_0$ as a vertex, can be uniquely determined by $k$ adjacent vertices  of $\bv_0$ along different dimensions in the grid. Along the $i$-th dimension $\bv_0$  has two adjacent vertices $(x_1, x_2, \ldots, x_i\pm1, \ldots, x_d)$, along positive and negative directions. 
        
        Now to count the number of $k$-cubes, that $\bv_0$ is a part of, we simply count the number of ways we can choose $k$ possible directions from the total $d$ directions, which is $\binom{d}{k}$. Then for each of these chosen directions, we can either choose the adjacent vertices along the positive or negative direction, i.e in $2^k$ ways.
        So the total number of $k$-cubes, that $\bv_0$ is a part of, is $2^k\binom{d}{k}$. Summing up over all dimensions we get:
        \begin{equation*}
            \sum_{k=0}^{d}\binom{d}{k}2^k = 3^d.
        \end{equation*}
        
        Note that we performed the calculation for an interior vertex. For the vertices on the boundary of the grid, each will be a part of fewer cubes. So we can bound the number of cubes, that a single vertex is a part of, by $3^d$.
    \end{proof}

    \begin{thm}
    \label{thm: EC_1}
        Let $I$ and $I^*$ be two $d$-dimensional binary images with vertex set $V$ s.t. they differ only at one voxel. Then along an arbitrary direction $\vec{u}$, 
        $$D(\EC_{\vec{u}, I}, \EC_{\vec{u}, I^*}) \leq 3^dn/\sqrt{d}$$ 
        where $n = |V|$ and 
        $D(\EC_{\vec{u}, I}, \EC_{\vec{u}, I^*})$ is the $l_2$-norm between $\EC_{\vec{u}, I}$ and $\EC_{\vec{u}, I^*}$, i.e.
        $$D(\EC_{\vec{u}, I}, \EC_{\vec{u}, I^*}) = \sqrt{\int_{h_{min}}^{h_{max}}(\chi(C_{\vec{u}, h}) - \chi(C^*_{\vec{u}, h}))^2dh}.$$
    \end{thm}
    
    \begin{proof}
    Let $C$ be the cubical complex with vertices $\bv_1,\,\bv_2, \ldots, \bv_n$ associated with image $I$, using the construction described in Section~\ref{subsec:CubicalComplex}. 
        Let  $h_1\leq h_2\leq\ldots\leq h_n$ be the ordered list of heights of the vertices 
        along $\vec{u}$. Then 
        $$D(\EC_{\vec{u}, I}, \EC_{\vec{u}, I^*})=\sqrt{\sum_{i=1}^{n-1}\int_{h_i}^{h_{i+1}}(\chi(C_{\vec{u}, h}) - \chi(C^*_{\vec{u}, h}))^2dh}.$$ 
        Since $EC$ is a piecewise constant function that changes only at the heights of vertices, we can rewrite it as,
        $$D(\EC_{\vec{u}, I}, \EC_{\vec{u}, I^*})=\sqrt{\sum_{i = 1}^{n-1}(h_{i+1} - h_{i})(\chi(C_{\vec{u}, h_i}) - \chi(C^*_{\vec{u}, h_i}))^2}.$$

        \noindent
       Let, $e=\max\{h_2-h_1, \ldots, h_n - h_{n-1}\}$. Then
        
        $$D(\EC_{\vec{u}, I}, \EC_{\vec{u}, I^*})\leq \sqrt{e}\sqrt{\sum_{i = 1}^{n-1}(\chi(C_{\vec{u}, h_i}) - \chi(C^*_{\vec{u}, h_i}))^2}.$$
         
        \noindent
        Since for $\x\in \R^n$, $\|\x\|_2 \leq \|\x\|_1$, we have 
        \begin{align*}
            D(\EC_{\vec{u}, I},& \EC_{\vec{u}, I^*}) \leq \sqrt{e}\sum_{i=1}^{n-1}|\chi(C_{\vec{u}, h_i}) - \chi(C^*_{\vec{u}, h_i})|\\
            &= \sqrt{e}\sum_{i=1}^{n-1}\left |\sum_{j=0}^{d}(-1)^j(\mathrm{Card}(C_{u, h_i}^j) - \mathrm{Card}(C_{u, h_i}^{*j}))\right |\\
            & \leq \sqrt{e}\sum_{i=1}^{n-1}\sum_{j=0}^{d}\left |(-1)^j(\mathrm{Card}(C_{u, h_i}^j) - \mathrm{Card}(C_{u, h_i}^{*j}))\right |
        \end{align*}
        Now for any sub-complex of $C$, the only cubes that can change are the ones that have $\bv_0$ as a constituent vertex. So, using Lemma \ref{lem:cubeCount}, we can bound the inner summation by $3^d$. Thus we have
        $$D(\EC_{\vec{u}, I}, \EC_{\vec{u}, I^*})\leq \sqrt{e}3^dn.$$
        
        Next, we provide a bound for $e$ to complete our proof. Every vertex $\bv_0=(x_1, \ldots, x_d)$ has at least $d$ adjacent vertices, say $\{\bv_i: i=1, \ldots, d\}$ where $\bv_i=(x_1, \ldots, x_i\pm 1, \ldots x_d)$.  
        We seek to find an upper bound of the minimum difference between the heights of the vertex $\bv_0$ and any of its adjacent vertices over all possible directions $\vec{u}=(u_1,\ldots, u_d)\in \bS^{d-1}$. This can be obtained by solving the following optimization problem:
        
        $$\displaystyle\max_{\vec{u} \in \bS^{d-1}}\min_{i \in \{1, 2, \ldots, d\}}(|\bv_i\cdot \vec{u} - \bv_0\cdot \vec{u}|)=\displaystyle\max_{\vec{u} \in \bS^{d-1}}\min_{i \in \{1, 2, \ldots, d\}} |u_i|$$ 
        with $\|\vec{u}\|=1$. The direction vector $\vec{u}$ that maximises this function is the vector with all equal components, i.e., $(1/\sqrt{d}, \ldots, 1/\sqrt{d})$. Thus, we obtain an upper bound of $e$ as $1/\sqrt{d}$.
    \end{proof}

    \begin{thm}
    Let $I$ and $I^*$ be two $d$-dimensional binary images with vertex set $V$ which differ at $k$ voxels $\bv_1, \ldots, \bv_k$. Then along an arbitrary direction 
    $\vec{u}$, $$D(\EC_{\vec{u}, I}, \EC_{\vec{u}, I^*}) \leq k3^dn/\sqrt{d}$$ where $n = |V|$.
    \label{thm:EC_full}
    \end{thm}
    
\begin{proof}
    From $I$, we construct a sequence of $k$ images $I_0, I_1, \ldots, I_k$, defined as follows:
    \begin{equation*}
        I_{i}(\bv) = \left\{
        \begin{array}{ll}
            I^*(\bv), & \bv = \bv_i \\
            I_{i-1}(\bv), & \text{ otherwise}
        \end{array}
    \right.
\end{equation*} 
for $i=1, 2, \ldots, k$ and $I_0 = I$. Observe that $I_k = I^*$ and that $I_i$ and $I_{i+1}$ differ by only one voxel for all $i$ from $0$ to $k-1$. Using the triangle inequality of a metric repeatedly and using Theorem \ref{thm: EC_1},
\begin{align*}
D(\EC_{\vec{u}, I}, \EC_{\vec{u}, I^*}) &\leq \sum_{i=0}^{k-1}D(\EC_{\vec{u}, I_i}, \EC_{\vec{u}, I_{i+1}})\\
& \leq \sum_{i=0}^{k-1}3^dn/\sqrt{d} = k3^dn/\sqrt{d}.
\end{align*}  
\end{proof}

\begin{cor}
    Let $I$ and $I^*$ be two $d$-dimensional binary images with vertex set $V$ s.t. they differ at $k$ voxels $\bv_1, \ldots, \bv_k$. Then, $$D(\ECT_{I}, \ECT_{I^*}) \leq k3^dn/\sqrt{d}\times \text{Surface area of }\bS^{d-1}$$
    where $n = |V|$.
\label{cor:ECT}
\end{cor}
\begin{proof}
    From theorem \ref{thm:EC_full}),  the distance between two $ECT$s, in equation (\ref{eqn:distance}), can be bounded as 
    $$D(\ECT_{I}, \ECT_{I^*}) \leq k3^dn/\sqrt{d} \times \displaystyle\int_{\vec{u}\in \bS^{d-1}} 1 \, du $$
    
\end{proof} 
    We note, in our proposed algorithm, we perform a thresholding operation on the real-valued image to convert it into a set of binary images. On performing this operation, our transformation ceases to be continuous. As a result, obtaining a similar upper bound on our ECT-based loss function is not possible. 
  
\begin{table*}[h]
\scriptsize
\begin{subtable}[h]{0.33\textwidth}
\label{table:IoU}
        \centering
\begin{tabular}{lll}
 \toprule
 \multicolumn{3}{c}{IoU Error $(\downarrow$)} \\
 \midrule
 Dataset & RBC &  Nuclei\\
 \midrule
 Baseline  &  $0.49 \pm 0.09$  &  0.64 $\pm$ 0.10\\
 Wasserstein &  $0.49 \pm 0.10$  & \textbf{0.62 $\pm$ 0.10}\\
 ECT & \textbf{0.48 $\pm$ 0.12} & 0.66 $\pm$ 0.11\\
 \bottomrule
\end{tabular}
\caption{}
\end{subtable}
\begin{subtable}[h]{0.33\textwidth}
        \centering
        \begin{tabular}{lll}
 \toprule
 \multicolumn{3}{c}{Volume Error~($\downarrow$)} \\
 \midrule
 Dataset & RBC &  Nuclei\\
 \midrule
 Baseline  &  $0.50 \pm 0.35$  &  0.61 $\pm$  0.57\\
 Wasserstein &  $0.45 \pm 0.35$  & 0.66 $\pm$ 0.58\\
 ECT & \textbf{0.40 $\pm$ 0.32} & \textbf{0.53 $\pm$ 0.51}\\
 \bottomrule
\end{tabular}
\caption{}
\end{subtable}
\begin{subtable}[h]{0.33\textwidth}
\centering
\begin{tabular}{lll}
 \toprule
 \multicolumn{3}{c}{Surface Error~($\downarrow$)} \\
 \midrule
 Dataset & RBC &  Nuclei\\
 \midrule
 Baseline  & 0.20 $\pm$ 0.14  &  0.38 $\pm$ 0.31\\
 Wasserstein &  0.22 $\pm$ 0.16  & 0.38 $\pm$ 0.31\\
 ECT & \textbf{0.19 $\pm$ 0.15} & \textbf{0.34 $\pm$ 0.29}\\
 \bottomrule
\end{tabular}
\caption{}
\end{subtable}
\caption{Performance of different variants of the SHAPR model on the Red blood cells and Nuclei dataset. Represented as mean $\pm$ standard deviation, with the best performing algorithm highlighted in \textbf{bold}.}
\label{table:errors}
\end{table*}

\section{Experimental Results}
\label{sec:Experiments}
We test the efficacy of our topological loss function by adding it to the SHAPR model and testing it on two biomedical datasets which have been used in the prior work \cite{Shapr, topo_shapr}. 
\begin{enumerate}
    \item Red Blood Cells(RBC): This is a dataset of 825 3D images obtained from a confocal microscope \cite{simionato2021red}. These cells are categorized into 9 designated categories: spherocytes, stomatocytes, discocytes, echinocytes, keratocytes, knizocytes, acanthocytes, cell clusters, and multilobates. The dimensionality of each image is $64\times64\times64$.
    \item Nuclei: This is a dataset of 887 3D images of nuclei of human-induced pluripotent stem cells. The dimensionality of each image is $64\times64\times64$.
\end{enumerate}
These datasets are publicly available.\footnote{https://hmgubox2.helmholtz-muenchen.de/index.php/s/YAds7dA2TcxSDtr}
Due to the limited dataset size, we follow the evaluation procedure of Waibel \etal \cite{topo_shapr}. That is we perform 5-fold cross-validation partitioning the dataset into five folds with a train/validation/test split of $60\%/20\%/20\%$. We ensure that each image of a dataset appears in the test split exactly once. We compare three different approaches to determine the improvements of our proposed loss. Namely, the baseline SHAPR \cite{Shapr}, SHAPR with the Wasserstein-based loss \cite{topo_shapr}, and finally SHAPR with our ECT-based loss. For the baseline SHAPR and the Wasserstein loss based implementation, we use the code made available by Waibel \etal \cite{topo_shapr}.\footnote{https://github.com/aidos-lab/SHAPR\_torch}

\begin{figure}[!h]
\begin{subfigure}{0.20\textwidth}
  \centering
  \includegraphics[width=\textwidth]{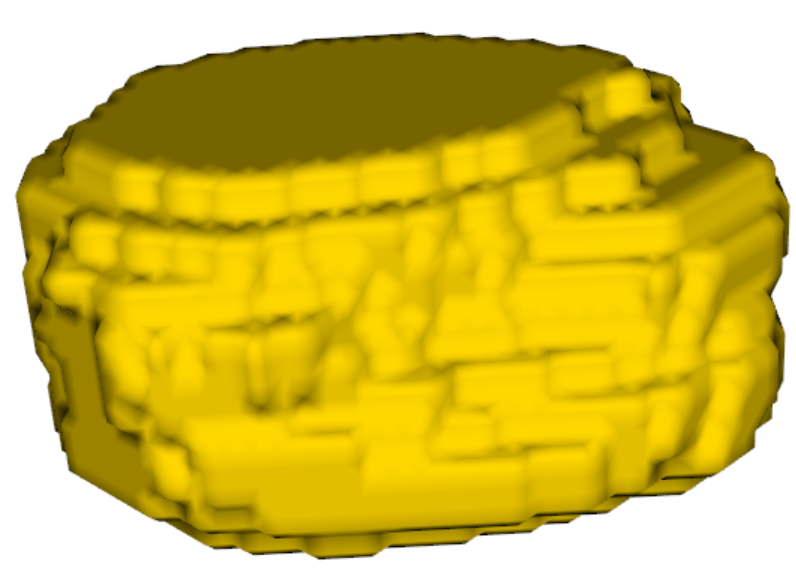}    
  \caption{Ground Truth}
\end{subfigure}%
\hfill
\begin{subfigure}{0.20\textwidth}
  \centering
  \includegraphics[width=\textwidth]{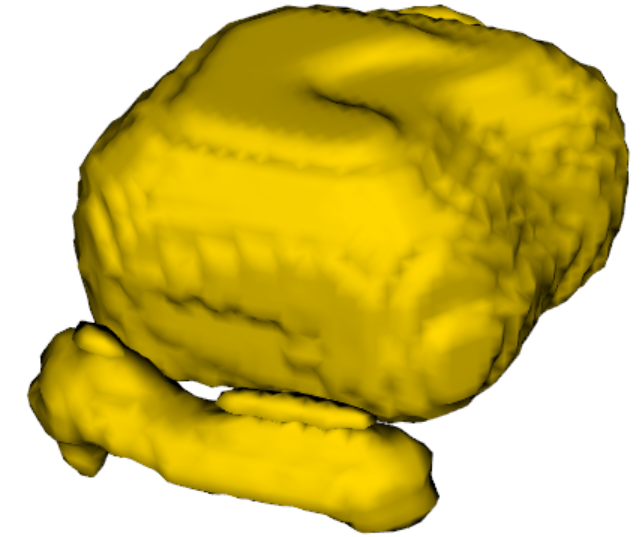}    
  \caption{SHAPR Baseline}
\end{subfigure}%
\hfill
\begin{subfigure}{0.20\textwidth}
  \centering
  \includegraphics[width=\textwidth]{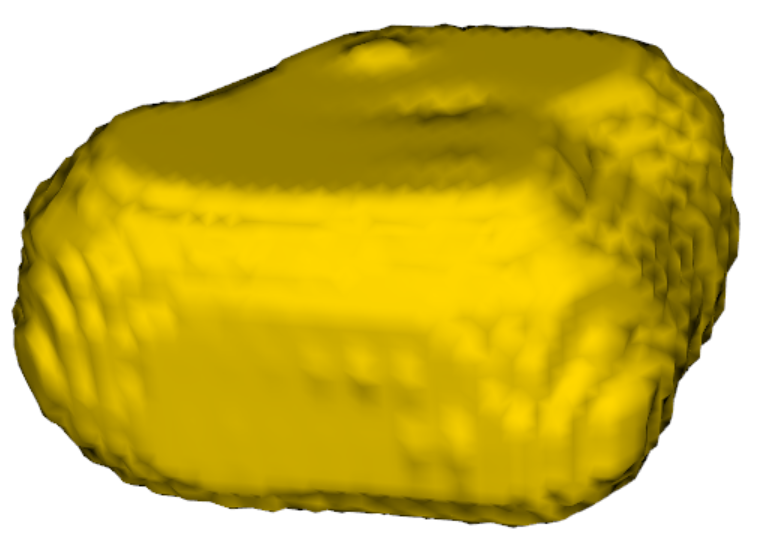}    
  \caption{Wasserstein}
\end{subfigure}%
\hfill
\begin{subfigure}{0.20\textwidth}
  \centering
  \includegraphics[width=\textwidth]{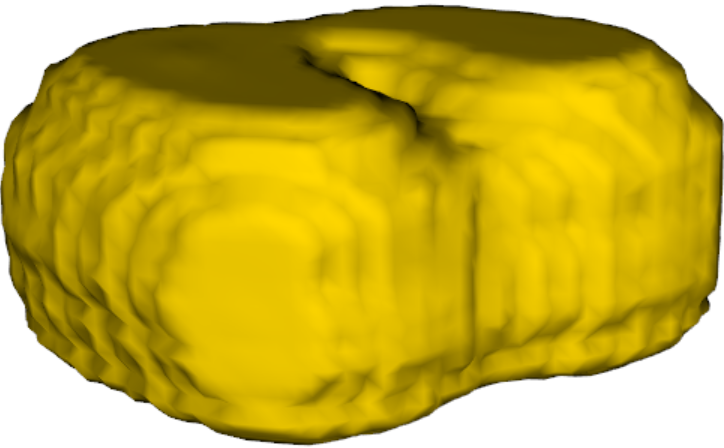}    
  \caption{ECT}
\end{subfigure}%
\caption{Qualitative results on the Nuclei dataset.  The result using  (b) SHAPR has topological artifacts, whereas, the results using topology-based loss functions (c) Wasserstein and (d) ECT do not have any topological artifacts. }
\label{fig:Nuclei}
\end{figure}

We follow the same training procedure as in Waibel \etal \cite{topo_shapr}, that is, we train all the variants of SHAPR for a maximum of 100 epochs, using early stopping with a patience parameter of 15 epochs. We also perform data augmentation before training by performing random horizontal or vertical flipping as well as $90^{\circ}$ rotations with a $33\%$ probability for an augmentation to be applied on a sample. We track our experiments using WANDB \cite{wandb}. 
In the testing phase, we apply Otsu's thresholding \cite{otsu} to convert our image into a binary image. This binary image is then compared with the ground truth to calculate three metrics from the prior works, namely, IoU error, relative Volume error and relative surface error. We drop the roughness error from the prior works \cite{Shapr, topo_shapr} as we believe it does not serve as a useful metric to measure the accuracy of the reconstruction. It is defined as the difference between the predicted image and a 3D smoothened Gaussian version of the image. As seen in Figure \ref{fig:Nuclei}, even the ground truth is rough in nature and will have a large roughness error. 

We train the baseline and Wasserstein loss based model using the hyperparameters reported in Waibel \etal\cite{topo_shapr}. For our ECT-based loss model, we use a scaling parameter $\lambda=0.01$. The number of thresholds we consider per pair of images is $40$ ($n$ in Algorithm \ref{algo:SHAPRAlgo}). The number of directions we consider in evaluating the integral of the distance function is 100 ($l$ in Algorithm \ref{algo:ECTAlgo}). Finally, the parameter $M$ or number of steps (Algorithm \ref{algo:EulerCurve}) we take as 30.  

We can see the results of our experiments in Table \ref{table:errors}. We observe that on most metrics over both the datasets the ECT-based loss performs the best. We see the most significant improvements in the Volume Error where the ECT-based loss outperforms the previous best by $11.2\%$ on the RBC dataset and $19.6\%$ on the Nuclei dataset. We also see significant improvements in the Surface Error, the ECT-based loss outperforms the previous best by $14.5\%$ on the RBC dataset and $10.8\%$ on the Nuclei dataset. 

\begin{figure}[!h]
\begin{subfigure}{0.20\textwidth}
  \centering
  \includegraphics[width=\textwidth]{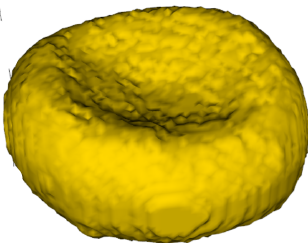}
  \caption{Ground Truth}
\end{subfigure}%
\hfill
\begin{subfigure}{0.20\textwidth}
  \centering
  \includegraphics[width=\textwidth]{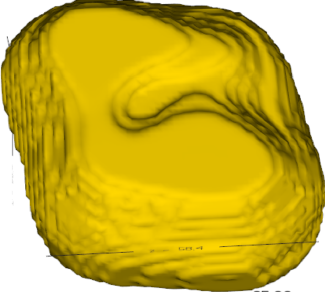}    
  \caption{SHAPR Baseline}

\end{subfigure}%
\hfill
\begin{subfigure}{0.20\textwidth}
  \centering
  \includegraphics[width=\textwidth]{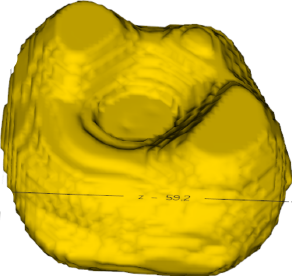}    
  \caption{Wasserstein}
\end{subfigure}%
\hfill
\begin{subfigure}{0.20\textwidth}
  \centering
  \includegraphics[width=\textwidth]{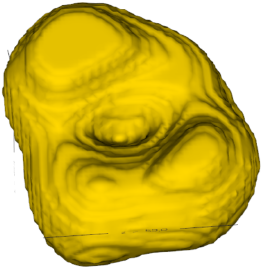}    
  \caption{ECT}
\end{subfigure}%
\caption{Qualitative results on the RBC dataset. The curvature of the valley in the ground truth is approximately captured by the topology-based loss functions (c) Wasserstein and (d) ECT, whereas, (b) the baseline is unable to capture it. }
\label{fig:RBC}
\end{figure}

We can visualize the outputs of the various methods on the Nuclei and RBC dataset in Figures \ref{fig:Nuclei} and \ref{fig:RBC}. It is interesting to observe that adding a topology based loss clearly improves the quality of the reconstruction, as neither of the topology based methods produce artifacts while the baseline does (Figure \ref{fig:Nuclei}).  This is expected since the topology based losses optimize for topological invariants to obtain topologically correct reconstructions. We also observe in Figure \ref{fig:RBC} that the topology based methods are trying to capture the valley in the ground truth, while the baseline does not. Note that in the current reconstruction problem we 
cannot expect perfect reconstructions since the problem is ill-posed.


\section{Conclusion}
\label{sec:conclusion}
In this paper, we present a novel ECT-based topological loss function that can be used to aid the training of neural networks for the challenging task of 3D image reconstruction from a single image. We not only show empirical improvement but also discuss some important theoretical properties of our loss and ECT in general. Our ECT-based loss can be used to describe the topological distance between any two images. Our method could thus potentially be employed to  aid neural networks in \emph{any} vision task, including image segmentation or 3D image reconstruction from multiple images.
Another natural extension of our work would be to consider the persistent homology transform (PHT) instead of the ECT. While both are injective, the persistence diagram is more informative than the Euler Curve
at the cost of additional computation. 
It would be interesting to explore whether this   provides any benefit.

\clearpage
{\small
\bibliographystyle{ieee_fullname}

\begin{thebibliography}{10}\itemsep=-1pt

\bibitem{allili2001cubical}
Madjid Allili, Konstantin Mischaikow, and Allen Tannenbaum.
\newblock Cubical homology and the topological classification of 2d and 3d
  imagery.
\newblock In {\em Proceedings 2001 international conference on image processing
  (Cat. No. 01CH37205)}, volume~2, pages 173--176. IEEE, 2001.

\bibitem{wandb}
Lukas Biewald.
\newblock Experiment tracking with weights and biases, 2020.
\newblock Software available from wandb.com.

\bibitem{chang2015shapenet}
Angel~X Chang, Thomas Funkhouser, Leonidas Guibas, Pat Hanrahan, Qixing Huang,
  Zimo Li, Silvio Savarese, Manolis Savva, Shuran Song, Hao Su, et~al.
\newblock Shapenet: An information-rich 3d model repository.
\newblock {\em arXiv preprint arXiv:1512.03012}, 2015.

\bibitem{cohen2005stability}
David Cohen-Steiner, Herbert Edelsbrunner, and John Harer.
\newblock Stability of persistence diagrams.
\newblock In {\em Proceedings of the twenty-first annual symposium on
  Computational geometry}, pages 263--271, 2005.

\bibitem{harer}
Herbert Edelsbrunner and John~L Harer.
\newblock {\em Computational topology: an introduction}.
\newblock American Mathematical Society, 2022.

\bibitem{3DRefPaper}
Xian-Feng Han, Hamid Laga, and Mohammed Bennamoun.
\newblock Image-based 3d object reconstruction: State-of-the-art and trends in
  the deep learning era.
\newblock {\em IEEE transactions on pattern analysis and machine intelligence},
  43(5):1578--1604, 2019.

\bibitem{hensel2021survey}
Felix Hensel, Michael Moor, and Bastian Rieck.
\newblock A survey of topological machine learning methods.
\newblock {\em Frontiers in Artificial Intelligence}, 4:681108, 2021.

\bibitem{hu2021topology}
Xiaoling Hu, Yusu Wang, Li Fuxin, Dimitris Samaras, and Chao Chen.
\newblock Topology-aware segmentation using discrete morse theory.
\newblock {\em arXiv preprint arXiv:2103.09992}, 2021.

\bibitem{WECT}
Qitong Jiang, Sebastian Kurtek, and Tom Needham.
\newblock The weighted euler curve transform for shape and image analysis.
\newblock In {\em Proceedings of the IEEE/CVF Conference on Computer Vision and
  Pattern Recognition Workshops}, pages 844--845, 2020.

\bibitem{kolotouros2019convolutional}
Nikos Kolotouros, Georgios Pavlakos, and Kostas Daniilidis.
\newblock Convolutional mesh regression for single-image human shape
  reconstruction.
\newblock In {\em Proceedings of the IEEE/CVF Conference on Computer Vision and
  Pattern Recognition}, pages 4501--4510, 2019.

\bibitem{lenail2019nn}
Alexander LeNail.
\newblock Nn-svg: Publication-ready neural network architecture schematics.
\newblock {\em J. Open Source Softw.}, 4(33):747, 2019.

\bibitem{thresholding}
Doruk Oner, Ad{\'e}lie Garin, Mateusz Kozinski, Kathryn Hess~Bellwald, and
  Pascal Fua.
\newblock Persistent homology with improved locality information for more
  effective delineation.
\newblock Technical report, 2022.

\bibitem{otsu}
Nobuyuki Otsu.
\newblock A threshold selection method from gray-level histograms.
\newblock {\em IEEE Transactions on Systems, Man, and Cybernetics},
  9(1):62--66, 1979.

\bibitem{fmri}
Bastian Rieck, Tristan Yates, Christian Bock, Karsten Borgwardt, Guy Wolf,
  Nicholas Turk-Browne, and Smita Krishnaswamy.
\newblock Uncovering the topology of time-varying {fMRI} data using cubical
  persistence.
\newblock {\em Advances in Neural Information Processing Systems},
  33:6900--6912, 2020.

\bibitem{simionato2021red}
Greta Simionato, Konrad Hinkelmann, Revaz Chachanidze, Paola Bianchi, Elisa
  Fermo, Richard van Wijk, Marc Leonetti, Christian Wagner, Lars Kaestner, and
  Stephan Quint.
\newblock Red blood cell phenotyping from 3d confocal images using artificial
  neural networks.
\newblock {\em PLoS Computational Biology}, 17(5):e1008934, 2021.

\bibitem{skraba2020wasserstein}
Primoz Skraba and Katharine Turner.
\newblock Wasserstein stability for persistence diagrams.
\newblock {\em arXiv preprint arXiv:2006.16824}, 2020.

\bibitem{sun2018pix3d}
Xingyuan Sun, Jiajun Wu, Xiuming Zhang, Zhoutong Zhang, Chengkai Zhang, Tianfan
  Xue, Joshua~B Tenenbaum, and William~T Freeman.
\newblock Pix3d: Dataset and methods for single-image 3d shape modeling.
\newblock In {\em Proceedings of the IEEE Conference on Computer Vision and
  Pattern Recognition}, pages 2974--2983, 2018.

\bibitem{turner2014persistent}
Katharine Turner, Sayan Mukherjee, and Doug~M Boyer.
\newblock Persistent homology transform for modeling shapes and surfaces.
\newblock {\em Information and Inference: A Journal of the IMA}, 3(4):310--344,
  2014.

\bibitem{topo_shapr}
Dominik J.~E. Waibel, Scott Atwell, Matthias Meier, Carsten Marr, and Bastian
  Rieck.
\newblock Capturing shape information with multi-scale topological loss terms
  for 3d reconstruction.
\newblock In Linwei Wang, Qi Dou, P.~Thomas Fletcher, Stefanie Speidel, and
  Shuo Li, editors, {\em Medical Image Computing and Computer Assisted
  Intervention~(MICCAI)}, pages 150--159, Cham, Switzerland, 2022. Springer.

\bibitem{Shapr}
Dominik J.~E. Waibel, Niklas Kiermeyer, Scott Atwell, Ario Sadafi, Matthias
  Meier, and Carsten Marr.
\newblock {SHAPR} predicts 3d cell shapes from 2d microscopic images.
\newblock {\em iScience}, 25(11):105298, 2022.

\bibitem{waibel2022diffusion}
Dominik J.~E. Waibel, Ernst R{\"o}ell, Bastian Rieck, Raja Giryes, and Carsten
  Marr.
\newblock A diffusion model predicts 3d shapes from 2d microscopy images.
\newblock {\em arXiv preprint arXiv:2208.14125}, 2022.

\bibitem{wang2020topogan}
Fan Wang, Huidong Liu, Dimitris Samaras, and Chao Chen.
\newblock Topogan: A topology-aware generative adversarial network.
\newblock In {\em Computer Vision--ECCV 2020: 16th European Conference,
  Glasgow, UK, August 23--28, 2020, Proceedings, Part III 16}, pages 118--136.
  Springer, 2020.

\end{thebibliography}

}

\end{document}